%% file: main.tex
\documentclass[10pt,twocolumn,a4paper]{article}

\usepackage{style/dlcc}
\dlccfinalcopy 

\usepackage{times}
\usepackage{epsfig}
\usepackage{graphicx}
\usepackage{amsmath}
\usepackage{amssymb}
\usepackage{float}

\usepackage[hidelinks,breaklinks=true,bookmarks=false]{hyperref}

\begin{document}

\title{StatPrompt-LLM: A Lightweight Statistical Refinement for Foundation-Style Time-Series Forecasting}

\author{Aniket Sinha\thanks{Both authors contributed equally to this work.}\\
{\tt\small aniket.sinha@sydney.edu.au} \\ 
The University of Sydney
\and
Veronica W. Tsou\footnotemark[1]\thanks{Corresponding author: veronica.tsou@sydney.edu.au}\\
{\tt\small veronica.tsou@sydney.edu.au} \\
The University of Sydney
}   

\maketitle

\input{sec/00_abstract}

\input{sec/01_introduction}

\input{sec/01.5_motivation}

\input{sec/02_methods}

\input{sec/03_experiments}

\input{sec/04_conclusion}

\section*{Acknowledgements}
The authors would like to thank Dr Hongwei Wen
for his valuable guidance in shaping the initial
research direction and for his constructive
feedback on the manuscript.

\clearpage
{\small
\bibliographystyle{style/ieee}
\bibliography{main}
}

\input{sec/appendix}

\end{document}

%% file: sec/00_abstract.tex
\begin{abstract}
   Time-series forecasting has recently seen a paradigm shift with the introduction of foundation-style models that adapt large pretrained language architectures to sequential data. However, effectively exposing fundamental temporal structures, such as trend, scale, and temporal dependence, to these models remains an open challenge, particularly when the available observations are limited. In this paper, we conduct an exploratory empirical study across different generations of forecasting architectures and propose StatPrompt-LLM, a lightweight statistical refinement of Time-LLM. StatPrompt-LLM augments the existing Time-LLM prompting mechanism with dataset-level statistical information and introduces a zero-initialised autoregressive bypass as a complementary linear forecasting pathway. We evaluate our framework on electricity and meteorological time-series benchmarks, including ETTh1, ETTh2, and Weather, under full-prefix and reduced-prefix configurations. Under the 10\% reduced-prefix configuration, StatPrompt-LLM achieved lower in-sample prediction errors than Time-LLM, with unweighted average relative reductions across datasets of 40.2\% in MSE and 21.4\% in MAE. Full-prefix results were more dataset-dependent. Ablation analyses revealed relatively modest and dataset-dependent changes when selected additional statistical descriptor groups were masked within the combined architecture. Computational profiling indicated additional model parameters and runtime overhead relative to Time-LLM.
\end{abstract}

%% file: sec/01_introduction.tex
\section{Introduction}
Time-series forecasting is a core problem in deep learning, with applications in energy systems, weather prediction, traffic management, finance, and other domains where future observations must be inferred from historical patterns. Real-world time series are difficult to model because they often contain a mixture of trend, seasonality, volatility, local fluctuations, and non-stationary shifts. Classical forecasting research has long emphasised the importance of trend, seasonality, and autocorrelation as central structures in temporal data~\cite{hyndman2021fpp3}. Large-scale forecasting competitions, such as the 4th Makridakis Competition (M4), have shown that strong forecasting performance often comes from effectively combining statistical structure with machine learning, rather than relying solely on increasingly complex machine learning models~\cite{makridakis2018m4}. These observations make forecasting a useful testbed for studying whether modern deep learning architectures can capture both simple statistical patterns and more complex temporal dependencies.

Recent forecasting research has progressed through several stages of model design. Simple linear models such as DLinear provide strong baselines by directly modelling trend-like and autoregressive structure~\cite{zeng2023dlinear}. Transformer-based models introduced attention-based sequence modelling for long-horizon forecasting, with later variants incorporating stronger time-series inductive biases such as decomposition and frequency-domain modelling~\cite{wu2021autoformer, zhou2022fedformer}. PatchTST further reformulates time series into patch-level tokens and applies channel-independent Transformer modelling~\cite{nie2023patchtst}. These developments reflect a broader shift from generic sequence modelling toward architectures designed to better capture temporal structure.

Despite this architectural progression, it remains unclear whether increasing model complexity consistently improves forecasting performance. Prior work has shown that simple linear baselines can remain competitive with more sophisticated Transformer models, suggesting that many benchmark datasets contain strong low-order statistical structure that may not always require large attention-based architectures~\cite{zeng2023dlinear}. This raises an important question: \textit{\textbf{How does augmenting a Time-LLM-based forecasting architecture with additional statistical descriptors and an autoregressive bypass affect its behaviour across different datasets and data configurations?}}

More recently, time-series forecasting has been influenced by the broader rise of foundation models. Models such as GPT4TS and Time-LLM adapt pretrained language models to time-series forecasting by aligning numerical sequences with language-model representations~\cite{zhou2023gpt4ts, jin2024timellm}. In parallel, time-series-specific foundation models such as Chronos, MOMENT, and TimesFM investigate large-scale pretraining directly for temporal data, through approaches such as value tokenisation, masked reconstruction, and decoder-only forecasting pretraining~\cite{ansari2024chronos, goswami2024moment, das2024timesfm}. These models suggest that pretrained sequence representations may improve generalisation and simplify forecasting pipelines, especially under zero-shot or data-limited settings. However, they also raise the question of how best to expose classical time-series structure, such as trend, volatility, periodicity, and non-stationarity, to large pretrained models.

A related development is the use of textual and multimodal context for time-series forecasting. Time-VLM extends the foundation-model direction by combining temporal, visual, and textual representations through frozen vision-language models~\cite{zhong2025timevlm}. In this framework, temporal features encode sequential dynamics, visual transformations expose spatial-temporal patterns, and textual prompts provide statistical and semantic context. These developments motivate further exploration of whether additional statistical descriptions and complementary temporal modelling pathways can influence the behaviour of existing foundation-style forecasting architectures.

Motivated by this gap, we conduct an exploratory empirical study of representative time-series forecasting models across different stages of architectural development. We evaluate DLinear, PatchTST, Autoformer, and a foundation-style model, Time-LLM, across ETTh1, ETTh2, and Weather. To assess both predictive behaviour and practical trade-offs, we organise the evaluation into four complementary experiments: full-prefix forecasting, reduced-prefix forecasting, ablation analysis of the proposed refinement, and computational efficiency comparison. Rather than treating this comparison as a simple leaderboard, we use it to examine how different modelling assumptions behave across datasets and data configurations.

Building on this comparison, we introduce StatPrompt-LLM, a lightweight refinement of Time-LLM that augments the foundation-style baseline with explicit time-series statistical descriptors. StatPrompt-LLM extracts a compact dataset-level statistical descriptor vector containing distributional summaries, trend and volatility measures, short-range temporal dependence, and stability information. These descriptors are used in two complementary ways: first, to augment the dataset description used by the language-model backbone; and second, as inputs to an autoregressive statistical bypass that maps the historical sequence and descriptor vector directly to the forecast horizon. The bypass is zero-initialised so that it contributes no additional output at the start of training, while the enhanced model can learn complementary temporal relationships through this pathway.

The contributions of this study are threefold. First, we provide a comparative evaluation of representative forecasting models, linking simple linear baselines, Transformer-based models, and foundation-style forecasting models within full-prefix, reduced-prefix, and efficiency-oriented configurations. Second, we present StatPrompt-LLM, a lightweight statistically augmented refinement of Time-LLM that introduces additional dataset-level statistical descriptions and an autoregressive bypass without redesigning the underlying foundation architecture. Third, we analyse the refinement through ablation experiments that mask selected additional temporal descriptor groups. The results show lower in-sample errors for the combined refinement under the reduced-prefix configuration, while full-prefix comparisons exhibit more mixed patterns across datasets. These findings characterise the behaviour of the combined refinement under the implemented protocol and do not establish generalisation to unseen observations.

%% file: sec/01.5_motivation.tex
\section{Problem Formulation}
\label{sec:problem_formulation}

To investigate how explicit statistical information can complement foundation-style forecasting models, we consider a multivariate time-series forecasting task. Given a historical context window of length $L$ with $M$ variables, the input sequence is denoted as
\begin{equation}
       X = \{x_1, x_2, \dots, x_L\}, 
    \quad x_t \in \mathbb{R}^{M},
\end{equation}
or equivalently $X \in \mathbb{R}^{L \times M}$. The objective is to predict the next $H$ time steps,
\begin{equation}
    \hat{Y} = \{\hat{y}_{L+1}, \hat{y}_{L+2}, \dots, \hat{y}_{L+H}\},
    \quad \hat{y}_t \in \mathbb{R}^{M},
\end{equation}
where $\hat{Y} \in \mathbb{R}^{H \times M}$ is the model prediction. The corresponding ground-truth future sequence is
\begin{equation}
    Y = \{y_{L+1}, y_{L+2}, \dots, y_{L+H}\},
    \quad Y \in \mathbb{R}^{H \times M}.
\end{equation}

The learning task is to estimate a forecasting function $f_{\theta}$ parameterised by $\theta$:
\begin{equation}
    \hat{Y} = f_{\theta}(X),
\end{equation}
where $f_{\theta}$ maps the historical multivariate context to the future multivariate trajectory. In this study, all experiments use a fixed context length of $L=96$ and forecasting horizon of $H=96$.

This formulation requires forecasting models to capture both low-order statistical structure, such as scale, trend, volatility, and local autoregressive dependence, and more complex nonlinear temporal patterns. Based on this task definition, we investigate three questions: (i) how representative linear, Transformer-based, and foundation-style models compare under a unified multivariate forecasting setting; (ii) how the combined statistical augmentation and autoregressive bypass affect prediction errors across the evaluated data configurations; and (iii) how sensitive the combined model is to masking selected groups of additional statistical descriptors. The methodology and experiments below address these questions across full-prefix, reduced-prefix, ablation, and efficiency settings.

%% file: sec/02_methods.tex
\section{Methodology}

Following the forecasting task defined in Section~\ref{sec:problem_formulation}, we describe the baseline forecasting models and introduce the proposed StatPrompt-LLM refinement. The method is designed to examine whether additional statistical descriptions and a direct autoregressive pathway can complement a Time-LLM-style forecasting model.

\subsection{Original Baselines and Foundation Framework}
The evaluated model implementations are sourced from the public ICML 2025 Time-VLM repository~\cite{zhong2025timevlm}. This repository provides the model classes used by our experimental pipeline, including DLinear, PatchTST, Autoformer, and Time-LLM; the multimodal Time-VLM architecture itself is not used in StatPrompt-LLM.

\begin{itemize}
    \item \textbf{Simple Linear Baseline (DLinear):} DLinear serves as a simple linear forecasting baseline, providing a reference for comparison with more complex architectures. It decomposes raw time-series data into trend and remainder components using a moving-average kernel and applies a linear layer to each component: $\hat{Y} = W_{\text{trend}} X_{\text{trend}} + W_{\text{season}} X_{\text{season}}$.
    \item \textbf{Time-Series Transformers (Autoformer and PatchTST):} Autoformer replaces standard self-attention with series decomposition and an Auto-Correlation mechanism. PatchTST divides each time series into local subsequence patches and applies channel-independent Transformer modelling.
    \item \textbf{Foundation-Style Forecasting Model (Time-LLM):} We adopt Time-LLM with a frozen GPT-2 backbone comprising six transformer blocks and trainable time-series adaptation components. Historical sequences are divided into patches and passed through a reprogramming layer that aligns numerical patch representations with learned projections of the language model's vocabulary embeddings. Time-LLM also constructs input-dependent textual prompts containing window-specific statistics. The prompt embeddings and reprogrammed patch embeddings are concatenated before processing by the frozen GPT-2 backbone, and a trainable projection head maps the resulting representations to the forecast horizon.
\end{itemize}

All models use an input context length of $L=96$ and forecasting horizon of $H=96$ while retaining their respective architectural configurations. The experimental pipeline supplies zero-filled marker and decoder tensors where required by the imported model interfaces; calendar-derived timestamp covariates are not used.

\subsection{Proposed Refinement: StatPrompt-LLM}
Rather than developing a new forecasting architecture from scratch, we propose StatPrompt-LLM, a lightweight refinement built on the Time-LLM framework. The approach combines additional dataset-level statistical descriptions with an autoregressive residual pathway.

\begin{figure*}[t]
    \centering
    \includegraphics[width=\textwidth]{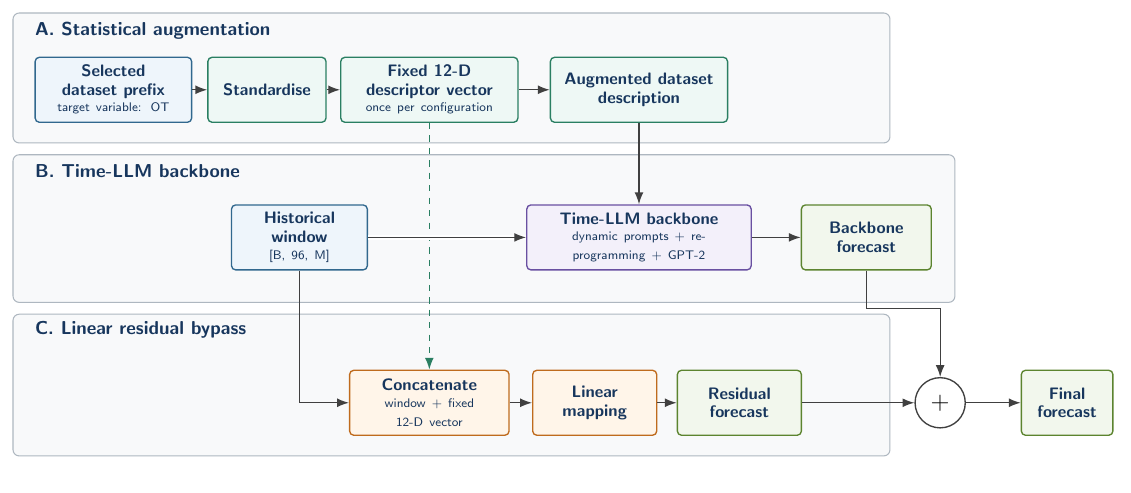}
    \caption{Overview of StatPrompt-LLM. A fixed descriptor vector is computed once from the standardised \texttt{OT} series of the selected dataset prefix. Ten descriptors augment the dataset description used by Time-LLM, while all twelve are supplied to a jointly multivariate linear residual bypass. Time-LLM's original window-specific prompts are retained, and the backbone and residual forecasts are added to obtain the final prediction.}
    \label{fig:statprompt_workflow}
\end{figure*}

Figure~\ref{fig:statprompt_workflow} illustrates the high-level workflow. A fixed descriptor vector is computed from the standardised \texttt{OT} series in the selected dataset prefix. Ten descriptor values augment the dataset description used by Time-LLM, while all twelve values are supplied to the linear residual bypass. The original Time-LLM window-specific statistical prompts remain active. The Time-LLM forecast and bypass output are added to obtain the final forecast.

\subsubsection{Statistical Prompt Augmentation}
Time-LLM already incorporates input-dependent statistical information into its textual prompts. Building on this mechanism, StatPrompt-LLM augments the dataset-level description with additional statistical summaries. A fixed 12-dimensional descriptor vector is extracted from the standardised target series within the selected dataset prefix, of which ten descriptors are formatted into the textual description. The minimum and maximum values are included in the numerical vector but are not repeated in this additional description. The original window-specific statistical prompts are retained. The resulting prompt embeddings are concatenated with reprogrammed time-series embeddings before being processed by the frozen language-model backbone. StatPrompt-LLM therefore changes the prompt content without introducing a new tokenisation, attention, or prompt-injection mechanism.

\subsubsection{Autoregressive Statistical Bypass}
To complement the Time-LLM backbone, we introduce an autoregressive statistical bypass that provides a direct linear mapping from historical observations and additional statistical descriptors to the forecast horizon. Let $X_{\text{flat}} \in \mathbb{R}^{B \times (L \cdot M)}$ denote the flattened, dataset-standardised input sequence and $P \in \mathbb{R}^{B \times 12}$ the descriptor matrix formed by repeating the configuration-level vector for each element in a batch. The bypass prediction is
\begin{equation}
    Y_{\text{bypass}} = [X_{\text{flat}} \oplus P] W_{\text{bypass}} + b_{\text{bypass}},
\end{equation}
where $\oplus$ denotes concatenation,
$[X_{\text{flat}} \oplus P] \in \mathbb{R}^{B \times ((L \cdot M)+12)}$,
$W_{\text{bypass}} \in \mathbb{R}^{((L \cdot M)+12) \times (H \cdot M)}$, and
$Y_{\text{bypass}} \in \mathbb{R}^{B \times (H \cdot M)}$. The jointly multivariate output is reshaped to the forecast dimensions and added to the Time-LLM prediction:
\begin{equation}
    Y_{\text{final}} = Y_{\text{LLM}} + Y_{\text{bypass}}.
\end{equation}
Because the descriptor vector is fixed within an experimental configuration, its contribution through this linear mapping is mathematically equivalent to an additional learned bias rather than window-specific statistical conditioning.

\textbf{Zero-initialisation strategy.}
The weights $W_{\text{bypass}}$ and biases $b_{\text{bypass}}$ are initialised to zero. Zero-initialisation ensures that the residual bypass contributes no additional output at the start of training. The bypass and the trainable Time-LLM adaptation components are subsequently optimised jointly.

\subsection{Dataset-Level Statistical Feature Extraction}
The additional statistical descriptors are computed once for each experimental configuration using the final variable (\texttt{OT}) of the selected dataset prefix. Let $z=(z_1,\ldots,z_N)$ denote this univariate prefix of length $N$. Before feature extraction, it is standardised using its mean and population standard deviation:
\begin{equation}
    z'_t = \frac{z_t - \mu_z}{\sigma_z + \epsilon},
\end{equation}
where $\epsilon=10^{-8}$ provides numerical stability. The resulting series is summarised using the following 12-dimensional descriptor vector:

\begin{enumerate}
    \item \textbf{Basic statistics (5):} mean, median, population standard deviation, minimum, and maximum of $z'$.
    \item \textbf{Trend (2):} the endpoint slope $(z'_N-z'_1)/(N+10^{-5})$ and its absolute value.
    \item \textbf{Volatility (2):} the mean of rolling sample variances computed with a window length of 10, and the global population variance of $z'$.
    \item \textbf{Short-range temporal dependence (1):} the Pearson lag-1 correlation between $(z'_1,\ldots,z'_{N-1})$ and $(z'_2,\ldots,z'_N)$.
    \item \textbf{Stability (2):} absolute differences in mean and population variance between the first and second halves of the selected prefix. The mean-shift descriptor is
    \begin{equation}
        \Delta_{\mu} = \left|\mu\left(z'_{1:\lfloor N/2 \rfloor}\right) - \mu\left(z'_{\lfloor N/2 \rfloor+1:N}\right)\right|.
    \end{equation}
\end{enumerate}

The prefix-level computation is separate from the per-window Time-LLM prompt statistics. Its cost grows with the selected prefix length and is incurred once when constructing a model for an experimental configuration. The dense bypass adds $(LM+13)HM$ parameters and corresponding matrix-multiplication costs, while the Time-LLM backbone retains its existing patching, reprogramming, tokenisation, and language-model operations.

%% file: sec/03_experiments.tex
\section{Experiments}
\subsection{Experimental Setup and Design}
\textbf{Datasets.}
We evaluate all models on three standard long-term time-series forecasting datasets: ETTh1, ETTh2, and Weather. ETTh1 and ETTh2 are electricity transformer temperature datasets, while Weather provides meteorological time-series observations and is treated as a 21-dimensional forecasting task in the adopted benchmark setting~\cite{zhong2025timevlm}. The benchmark datasets and model implementations are sourced from the public Time-VLM repository, while data preprocessing and execution are handled by our experimental pipeline.

\textbf{Models.}
We compare representative forecasting models across different stages of recent time-series model development. DLinear is included as a simple linear baseline, PatchTST as a modern patch-based Transformer model, and Autoformer as a decomposition-based Transformer baseline. We use Time-LLM as the foundation-style baseline and compare it with our proposed refinement, StatPrompt-LLM. These baselines are selected from model families considered in the Time-VLM benchmark~\cite{zhong2025timevlm}.

\textbf{Implementation.}
All models are evaluated on multivariate forecasting tasks using a fixed input context window of $L=96$ and prediction horizon of $H=96$. Models are trained using the Adam optimiser with a learning rate of $0.001$, batch size 32, and three training epochs. Predictions are evaluated using the final model state, without validation-based checkpoint selection.

\textbf{Metrics.}
Prediction error is evaluated using mean squared error (MSE) and mean absolute error (MAE), where lower values indicate smaller discrepancies between predictions and targets. MSE penalises larger errors more strongly, while MAE measures the average absolute deviation. Prediction errors are calculated on dataset-standardised values. For computational efficiency, we also report trainable parameters, total parameters, training time, and inference time in Appendix~\ref{sec:training and inference parameters}.

\textbf{Experimental design.}
We organise the evaluation into four experiments. 
\begin{itemize}
    \item \textbf{Experiment 1: Full-prefix forecasting.}
    All models are trained and evaluated using overlapping forecasting windows constructed from the complete available dataset. This provides an in-sample comparison under the full-prefix configuration.
    \item \textbf{Experiment 2: Reduced-prefix forecasting.}
    Models are trained and evaluated using windows constructed from the first 10\% of each dataset. This experiment explores model behaviour when a smaller portion of the available observations is used.
    \item \textbf{Experiment 3: StatPrompt-LLM ablation analysis.} 
    We evaluate StatPrompt-LLM variants under the 10\% reduced-prefix configuration by masking selected additional descriptor groups. Specifically, we compare the full model with variants that mask the added lag-1 Pearson correlation descriptor, mask trend and stability features, or retain only the five basic additional descriptors. The masks affect both the augmented dataset description and the descriptor coordinates supplied to the bypass, while the original Time-LLM window-specific prompts and the bypass architecture remain active.
    \item \textbf{Experiment 4: Computational efficiency analysis.} 
    We record trainable parameters, total parameters, training time, and inference time to assess the computational overhead introduced by StatPrompt-LLM relative to the original Time-LLM baseline and other forecasting models.
\end{itemize}

Models were evaluated on the same forecasting windows used during training, without a separate held-out test set. Accordingly, the reported errors are interpreted as in-sample comparisons rather than evidence of generalisation to unseen observations.

All experiments were run on an Apple Mac mini with M4 Pro chip and 64GB of RAM.

\subsection{Comparison of Results}\label{sec:comparison of results}
\subsubsection{Full-Prefix Forecasting}
Table~\ref{tab:full_data_results} reports the in-sample prediction errors under the full-prefix configuration. StatPrompt-LLM achieves lower MSE than the original Time-LLM baseline on ETTh1 and ETTh2. On ETTh1, MSE decreases from 0.4263 to 0.3597, corresponding to a relative difference of 15.6\%. On ETTh2, it decreases from 0.2804 to 0.2250, corresponding to 19.8\%. The MAE results are more mixed: StatPrompt-LLM achieves a lower error on ETTh1, a similar error on ETTh2, and a higher error on Weather.

The full-prefix results are not uniformly favourable to StatPrompt-LLM. On Weather, StatPrompt-LLM has higher errors than the original Time-LLM baseline, and PatchTST achieves the lowest errors. One possible explanation is that the Weather dataset contains patterns that are not well represented by the compact dataset-level descriptor vector. This interpretation remains speculative; the comparisons establish a dataset-dependent pattern rather than its cause.

\begin{table}[!htbp]
\centering
\begin{tabular}{llcc}
\hline
Dataset & Model & MSE & MAE \\
\hline
ETTh1 & DLinear & 0.3699 & 0.4128 \\
 & PatchTST & 0.3634 & \textbf{0.4108} \\
 & Autoformer & 0.5050 & 0.5023 \\
 & Time-LLM & 0.4263 & 0.4499 \\
 & StatPrompt-LLM & \textbf{0.3597} & 0.4177 \\
\hline
ETTh2 & DLinear & 0.2500 & 0.3012 \\
 & PatchTST & 0.2408 & \textbf{0.3006} \\
 & Autoformer & 0.2917 & 0.3427 \\
 & Time-LLM & 0.2804 & 0.3179 \\
 & StatPrompt-LLM & \textbf{0.2250} & 0.3191 \\
\hline
Weather & DLinear & 0.4871 & 0.3444 \\
 & PatchTST & \textbf{0.4381} & \textbf{0.3091} \\
 & Autoformer & 0.5328 & 0.4096 \\
 & Time-LLM & 0.5201 & 0.3599 \\
 & StatPrompt-LLM & 0.5586 & 0.4118 \\
\hline
\end{tabular}
\vspace{3pt}
\caption{In-sample prediction errors under the full-prefix configuration on ETTh1, ETTh2, and Weather. MSE denotes mean squared error and MAE denotes mean absolute error on dataset-standardised values; lower values indicate smaller errors. Bold values indicate the lowest error for each dataset and metric.}
\label{tab:full_data_results}
\end{table}

\subsubsection{Reduced-Prefix Forecasting}
Table~\ref{tab:few_shot_results} reports the in-sample errors for windows constructed from the first 10\% of each dataset. StatPrompt-LLM achieves the lowest MSE and MAE among the evaluated models across all three datasets. Relative to the original Time-LLM baseline, its MSE is lower by 39.9\% on ETTh1, 41.9\% on ETTh2, and 38.8\% on Weather. The corresponding MAE differences are 22.3\%, 23.4\%, and 18.4\%, respectively.

These results indicate that the combined StatPrompt-LLM refinement achieved lower in-sample prediction errors than the evaluated baselines under the reduced-prefix configuration. Although the additional statistical descriptions and autoregressive bypass provide complementary modelling pathways, their individual contributions cannot be determined from the current experiments. The observed differences should therefore not be interpreted as evidence of improved forecasting generalisation.

\begin{table}[!htbp]
\centering
\begin{tabular}{llcc}
\hline
Dataset & Model & MSE & MAE \\
\hline
ETTh1 & DLinear & 0.5881 & 0.5904 \\
 & PatchTST & 0.6191 & 0.5971 \\
 & Autoformer & 0.7423 & 0.6759 \\
 & Time-LLM & 0.6804 & 0.6320 \\
 & StatPrompt-LLM & \textbf{0.4086} & \textbf{0.4912} \\
\hline
ETTh2 & DLinear & 0.6192 & 0.5562 \\
 & PatchTST & 0.6347 & 0.5527 \\
 & Autoformer & 0.7445 & 0.6180 \\
 & Time-LLM & 0.7161 & 0.5905 \\
 & StatPrompt-LLM & \textbf{0.4164} & \textbf{0.4523} \\
\hline
Weather & DLinear & 0.5784 & 0.5020 \\
 & PatchTST & 0.5608 & 0.4677 \\
 & Autoformer & 0.6215 & 0.5431 \\
 & Time-LLM & 0.6212 & 0.5143 \\
 & StatPrompt-LLM & \textbf{0.3801} & \textbf{0.4196} \\
\hline
\end{tabular}
\vspace{3pt}
\caption{In-sample prediction errors using windows from the first 10\% of each available dataset. MSE and MAE are calculated on dataset-standardised values; lower values indicate smaller errors. Bold values indicate the lowest error for each dataset and metric.}
\label{tab:few_shot_results}
\end{table}

\subsubsection{Summary of Forecasting Performance}
To make the comparison between Time-LLM and StatPrompt-LLM more explicit, Table~\ref{tab:relative_improvement} summarises their relative differences. Positive values indicate that StatPrompt-LLM achieves lower error than Time-LLM, while negative values indicate worse performance.

Under the full-prefix configuration, StatPrompt-LLM achieves lower MSE on ETTh1 and ETTh2, but this pattern does not extend to Weather. Under the reduced-prefix configuration, it achieves lower MSE and MAE than Time-LLM across all evaluated datasets. The unweighted arithmetic mean of the three dataset-specific relative differences is 40.2\% for MSE and 21.4\% for MAE. These comparisons show that the combined refinement produced lower in-sample errors under the reduced-prefix configuration, while the full-prefix results were more dataset-dependent.

\begin{table}[!htbp]
\centering
\begin{tabular}{lcccc}
\hline
Dataset & \multicolumn{2}{c}{Full-prefix} & \multicolumn{2}{c}{Reduced-prefix} \\
 & MSE & MAE & MSE & MAE \\
\hline
ETTh1 & 15.6\% & 7.1\% & 39.9\% & 22.3\% \\
ETTh2 & 19.8\% & -0.4\% & 41.9\% & 23.4\% \\
Weather & -7.4\% & -14.4\% & 38.8\% & 18.4\% \\
\hline
Average & 9.3\% & -2.6\% & 40.2\% & 21.4\% \\
\hline
\end{tabular}
\vspace{3pt}
\caption{Relative differences between StatPrompt-LLM and Time-LLM. Positive values indicate lower in-sample error for StatPrompt-LLM; negative values indicate higher error. The Average row is the unweighted arithmetic mean of the three dataset-specific percentages.}
\label{tab:relative_improvement}
\end{table}

\subsection{Ablation Study of Additional Statistical Descriptors}
Table~\ref{tab:ablation_study_results} reports the results of masking selected additional statistical descriptor groups within StatPrompt-LLM. The variants generally exhibit modest but dataset-dependent differences relative to the full configuration. These observations characterise sensitivity to the added descriptors within the combined architecture, rather than isolating the contributions of prompting and the autoregressive bypass.

\begin{table}[!htbp]
\centering
\resizebox{\columnwidth}{!}{
\begin{tabular}{llcc}
\hline
Dataset & Model & MSE & MAE \\
\hline
ETTh1 & Time-LLM & 0.6804 & 0.6320 \\
 & StatPrompt-LLM & 0.4086 & 0.4912 \\
 & StatPrompt-LLM (No Added Lag-1 Correlation) & 0.4067 & 0.4939 \\
 & StatPrompt-LLM (No Trend/Stability) & \textbf{0.3990} & \textbf{0.4849} \\
 & StatPrompt-LLM (Basic Only) & 0.4027 & 0.4869 \\
\hline
ETTh2 & Time-LLM & 0.7161 & 0.5905 \\
 & StatPrompt-LLM & 0.4164 & \textbf{0.4523} \\
  & StatPrompt-LLM (No Added Lag-1 Correlation) & 0.4159 & 0.4538 \\
 & StatPrompt-LLM (No Trend/Stability) & 0.4157 & 0.4540 \\
 & StatPrompt-LLM (Basic Only) & \textbf{0.4143} & 0.4526 \\
\hline
Weather & Time-LLM & 0.6212 & 0.5143 \\
 & StatPrompt-LLM & 0.3801 & 0.4196 \\
  & StatPrompt-LLM (No Added Lag-1 Correlation) & 0.3827 & \textbf{0.4184} \\
 & StatPrompt-LLM (No Trend/Stability) & 0.3967 & 0.4273 \\
 & StatPrompt-LLM (Basic Only) & \textbf{0.3797} & 0.4201 \\
\hline
\end{tabular}
}
\vspace{3pt}
\caption{Ablation study under the 10\% reduced-prefix configuration. The variants mask the added lag-1 correlation descriptor, mask added trend/stability descriptors, or retain only the five basic additional descriptors. All variants retain the original Time-LLM window-specific prompts and the autoregressive bypass. Errors are in dataset-standardised units.}
\label{tab:ablation_study_results}
\end{table}

The basic-only variant, which retains the five basic additional descriptors, remains close to the full StatPrompt-LLM configuration. Masking the added lag-1 correlation or trend/stability groups also produces comparatively small changes on ETTh1 and ETTh2, whereas removing trend/stability has a larger adverse effect on Weather MSE. These results indicate limited incremental effects from some added descriptor groups in the evaluated configurations; they do not show that the basic descriptors independently cause the overall difference from Time-LLM.

One possible explanation is that the autoregressive bypass already provides a direct pathway for modelling low-order temporal relationships. The original Time-LLM input-dependent prompts, including their window-specific trend and lag information, also remain present in every ablation variant. Consequently, the current ablations cannot isolate the effects of dataset-level prompt augmentation from those of the bypass or inherited prompting mechanism.

\subsection{Computational Efficiency}
The detailed timing and parameter records are reported in Appendix~\ref{sec:training and inference parameters}. To assess computational efficiency, we compare training and inference-loop times across six matched configurations covering the three datasets under full-prefix and reduced-prefix settings. Averaging these six configurations gives mean training-loop times of 1821.462 seconds for Time-LLM and 1875.873 seconds for StatPrompt-LLM, with corresponding inference-loop times of 233.224 and 248.432 seconds.

The matched comparisons indicate that StatPrompt-LLM introduces additional parameters and higher training-loop and inference-loop times relative to Time-LLM. The overhead varies across datasets and configurations, reflecting a computational trade-off associated with the combined refinement. These measurements are whole-run wall-clock measurements from the experimental implementation rather than controlled per-example latency benchmarks.

%% file: sec/04_conclusion.tex
\section{Conclusion}

In this study, we presented an exploratory comparison of contemporary time-series forecasting paradigms and introduced StatPrompt-LLM, a statistically augmented refinement of Time-LLM. The implementation combines dataset-level statistical prompt augmentation with a zero-initialised autoregressive residual pathway while retaining the frozen GPT-2 backbone and existing Time-LLM prompting mechanism.

The results indicate that the combined StatPrompt-LLM refinement achieved lower in-sample errors under the reduced-prefix configurations, while the full-prefix comparisons showed mixed performance across datasets. Ablation analyses show relatively modest but dataset-dependent changes when selected additional descriptor groups are masked. Because the original Time-LLM prompts and autoregressive bypass remain active in each variant, the contributions of prompt augmentation and the bypass cannot be isolated. Matched computational comparisons further show that the combined refinement adds parameters and incurs higher training-loop and inference-loop times than Time-LLM. These findings provide an initial characterisation of how dataset-level statistical augmentation and a linear residual pathway behave within the Time-LLM framework, while motivating further evaluation under independent testing protocols.

The evaluation was limited to three datasets, a single input length and forecasting horizon, and an in-sample assessment protocol. The available results reflect single runs, and the configuration-level descriptors are computed from the full selected prefix. Consequently, the comparisons do not establish generalisation to unseen observations. Further evaluation with independent test data would be needed to assess out-of-sample forecasting performance.

Future work should evaluate StatPrompt-LLM on independent held-out data, replicate the comparisons across multiple random seeds, isolate the contributions of prompt augmentation and the residual bypass, and investigate longer historical context windows.

%% file: sec/appendix.tex
\appendix
\section{Appendix}

The implementation is available at \url{https://github.com/greatoverlordgollum/statprompt-llm}.

The authors acknowledge the use of generative AI tools, including Gemini and ChatGPT/Codex, to assist with coding and implementation auditing. The authors reviewed and take responsibility for the final content.

\subsection{Parameters, Training and Inference Times}\label{sec:training and inference parameters}

For computational efficiency, we report trainable parameters, total parameters, training time and inference-loop time in Tables \ref{tab:full_data_metrics}, \ref{tab:few_shot_metrics}, and \ref{tab:ablation_study_metrics}. In these tables:

\begin{itemize}
    \item Trainable parameters refer to the specific weights and biases within the neural network that are actively updated and learned through gradient descent during the training phase.
    \item Total parameters refers to the absolute total number of all weights and biases stored in the model, including both the actively updated layers and any large, ``frozen" pre-trained layers (such as the LLM backbone) that remain static.
    \item Training time refers to the wall-clock duration of the three-epoch training loop measured by the experimental implementation.
    \item Inference time refers to the wall-clock duration of the subsequent evaluation loop over the same windows.
\end{itemize}


\begin{table*}[!htbp]
\centering
\resizebox{\textwidth}{!}{
\begin{tabular}{llcccc}
\hline
Dataset & Model & Trainable Parameters & Total Parameters & Training Time (s) & Inference Time (s) \\
\hline
ETTh1 & DLinear & 18624 & 18624 & 0.904 & 0.157 \\
 & PatchTST & 62944 & 62944 & 114.265 & 10.420 \\
 & Autoformer & 44103 & 44103 & 163.426 & 34.988 \\
 & Time-LLM & 51604400 & 133516976 & 990.343 & 130.751 \\
 & StatPrompt-LLM & 52064720 & 133977296 & 1007.313 & 132.112 \\
\hline
ETTh2 & DLinear & 18624 & 18624 & 0.874 & 0.125 \\
 & PatchTST & 62944 & 62944 & 125.551 & 10.336 \\
 & Autoformer & 44103 & 44103 & 164.304 & 35.005 \\
 & Time-LLM & 51604400 & 133516976 & 957.998 & 125.927 \\
 & StatPrompt-LLM & 52064720 & 133977296 & 978.703 & 132.152 \\
\hline
Weather & DLinear & 18624 & 18624 & 6.681 & 1.036 \\
 & PatchTST & 62944 & 62944 & 936.319 & 76.114 \\
 & Autoformer & 48597 & 48597 & 529.375 & 114.941 \\
 & Time-LLM & 51604400 & 133516976 & 8051.164 & 1021.793 \\
 & StatPrompt-LLM & 55694864 & 137607440 & 8304.849 & 1099.038 \\
\hline
\end{tabular}
}
\vspace{3pt}
\caption{Parameter and timing results under the full-prefix configuration. Times are whole-loop wall-clock measurements from the experimental implementation.}
\label{tab:full_data_metrics}
\end{table*}


\begin{table*}[!htbp]
\centering
\resizebox{\textwidth}{!}{
\begin{tabular}{llcccc}
\hline
Dataset & Model & Trainable Parameters & Total Parameters & Training Time (s) & Inference Time (s) \\
\hline
ETTh1 & DLinear & 18624 & 18624 & 0.0973 & 0.012 \\
 & PatchTST & 62944 & 62944 & 9.244 & 0.892 \\
 & Autoformer & 44103 & 44103 & 13.662 & 3.029 \\
 & Time-LLM & 51604400 & 133516976 & 92.369 & 11.841 \\
 & StatPrompt-LLM & 52064720 & 133977296 & 96.168 & 12.214 \\
\hline
ETTh2 & DLinear & 18624 & 18624 & 0.077 & 0.011 \\
 & PatchTST & 62944 & 62944 & 10.157 & 0.954 \\
 & Autoformer & 44103 & 44103 & 14.326 & 3.129 \\
 & Time-LLM & 51604400 & 133516976 & 86.285 & 11.062 \\
 & StatPrompt-LLM & 52064720 & 133977296 & 88.865 & 11.304 \\
\hline
Weather & DLinear & 18624 & 18624 & 0.659 & 0.104 \\
 & PatchTST & 62944 & 62944 & 88.323 & 6.877 \\
 & Autoformer & 48597 & 48597 & 49.109 & 10.911 \\
 & Time-LLM & 51604400 & 133516976 & 750.614 & 97.969 \\
 & StatPrompt-LLM & 55694864 & 137607440 & 779.341 & 103.770 \\
\hline
\end{tabular}
}
\vspace{3pt}
\caption{Parameter and timing results for windows constructed from the first 10\% of each available dataset. Times are whole-loop wall-clock measurements from the experimental implementation.}
\label{tab:few_shot_metrics}
\end{table*}


\begin{table*}[!htbp]
\centering
\resizebox{\textwidth}{!}{
\begin{tabular}{llcccc}
\hline
Dataset & Model & Trainable Parameters & Total Parameters & Training Time (s) & Inference Time (s) \\
\hline
ETTh1 & Time-LLM & 51604400 & 133516976 & 92.369 & 11.841 \\
 & StatPrompt-LLM & 52064720 & 133977296 & 96.168 & 12.214 \\
 & StatPrompt-LLM (No Added Lag-1 Correlation) & 52064720 & 133977296 & 92.809 & 11.622 \\
 & StatPrompt-LLM (No Trend/Stability) & 52064720 & 133977296 & 91.376 & 11.789 \\
 & StatPrompt-LLM (Basic Only) & 52064720 & 133977296 & 92.022 & 11.586 \\
\hline
ETTh2 & Time-LLM & 51604400 & 133516976 & 86.285 & 11.062 \\
 & StatPrompt-LLM & 52064720 & 133977296 & 88.865 & 11.304 \\
  & StatPrompt-LLM (No Added Lag-1 Correlation) & 52064720 & 133977296 & 90.671 & 11.397 \\
 & StatPrompt-LLM (No Trend/Stability) & 52064720 & 133977296 & 88.635 & 11.679 \\
 & StatPrompt-LLM (Basic Only) & 52064720 & 133977296 & 89.162 & 11.251 \\
\hline
Weather & Time-LLM & 51604400 & 133516976 & 750.614 & 97.969 \\
 & StatPrompt-LLM & 55694864 & 137607440 & 779.341 & 103.770 \\
  & StatPrompt-LLM (No Added Lag-1 Correlation) & 55694864 & 137607440 & 774.146 & 102.971 \\
 & StatPrompt-LLM (No Trend/Stability) & 55694864 & 137607440 & 803.710 & 106.174 \\
 & StatPrompt-LLM (Basic Only) & 55694864 & 137607440 & 786.939 & 104.048 \\
\hline
\end{tabular}
}
\vspace{3pt}
\caption{Parameter and timing results for the descriptor ablations under the 10\% reduced-prefix configuration.}
\label{tab:ablation_study_metrics}
\end{table*}